\documentclass[runningheads]{llncs}

\usepackage[year=2026]{eccv}

\usepackage{eccvabbrv}

\usepackage{graphicx}
\usepackage{booktabs}

\usepackage[accsupp]{axessibility}  

\usepackage{hyperref}

\usepackage{orcidlink}

\usepackage{xcolor}
\usepackage{zebra-goodies}
\usepackage{bm}
\usepackage{wrapfig}
\usepackage[width=122mm,left=12mm,paperwidth=146mm,height=193mm,top=12mm,paperheight=217mm]{geometry}

\begin{document}

\title{How good was my shot? \\ Quantifying Player Skill Level in Table Tennis} 


\author{Akihiro Kubota\orcidlink{0009-0007-3821-8119} \and
Tomoya Hasegawa \and
Ryo Kawahara\orcidlink{0000-0002-9819-3634} \and
Ko Nishino\orcidlink{0000-0002-3534-3447}}

\authorrunning{A.~Kubota et al.}

\institute{Graduate School of Informatics, Kyoto University, Kyoto, Japan
\url{https://vision.ist.i.kyoto-u.ac.jp/}}

\maketitle

\begin{figure}[]
\vspace{-24pt}
  \centering
  \includegraphics[width=\textwidth]{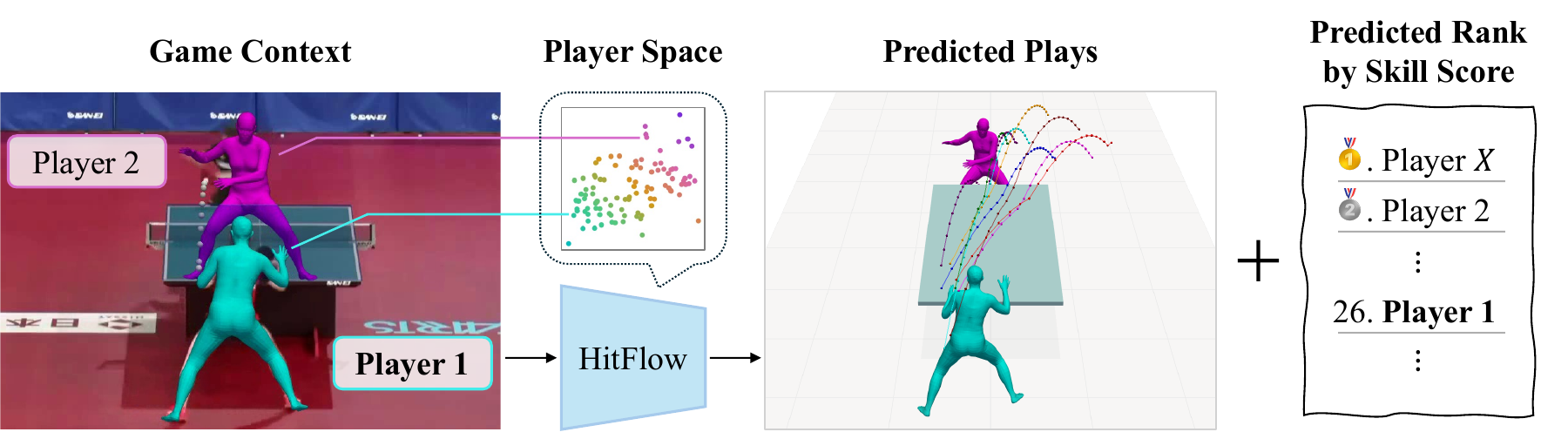}
  \caption{
We demonstrate skill level quantification using dyadic sports as a primary example. From match videos, we jointly learn player-specific generative models and a player space defined by their latent embeddings. We show that this learned player space contains rich information regarding individual proficiency, enabling the quantification of skill scores and both relative and absolute ranking predictions.
  }
  \label{fig:teaser}
\end{figure}
\vspace{-24pt}

\begin{abstract}
Gauging an individual’s skill level is crucial, as it inherently shapes their behavior. Quantifying skill, however, is challenging because it is latent to the observed actions. To explore skill understanding in human behavior, we focus on dyadic sports—specifically table tennis—where skill manifests not just in complex movements, but in the subtle nuances of execution conditioned on game context. Our key idea is to learn a generative model of each player’s tactical racket strokes and jointly embed them in a common latent space that encodes individual characteristics, including those pertaining to skill levels. By training these player models on a large-scale dataset of 3D-reconstructed professional matches and conditioning them on comprehensive game context—including player positioning and opponent behaviors—the models capture individual tactical identities within their latent space. We probe this learned player space and find that it reflects distinct play styles and attributes that collectively represent skill. By training a simple relative ranking network on these embeddings, we demonstrate that both relative and absolute skill predictions can be achieved. These results demonstrate that the learned player space effectively quantifies skill levels, providing a foundation for automated skill assessment in complex, interactive behaviors.
 
\end{abstract}

\section{Introduction} \label{sec:intro}
Judging another's skills is almost innate to our daily lives. At a restaurant, we decide whether to return by judging the cook's skill from their dishes. In a cake shop, we gauge the p\^{a}tissier's craft from the cakes on display. When seeing a doctor, we assess whether to seek a second opinion through our interactions. Judging skills is also essential to nurturing them---if you cannot evaluate a skill, you cannot teach it. 

Skills become more apparent when the craft relies on body movements. In soccer, we judge a player's level from the way they carry themselves in a crowded penalty zone. We connect with our favorite actor through the subtle body movements that express the complex emotions. One might associate skills with Olympic gymnasts performing otherworldly acts. Skills are, however, not just about defying the laws of physics. They are also about mundane movement such as moving a hand, but executed in the precise situation and manner. Consider a sushi chef commanding \$500 per meal. Their body movements are just as ordinary as yours cutting fish and rolling sushi rice, but their control of their body movements results in a heavenly, light textured sushi. Skills lie as much in the sequence and variation of the body movements as in the difficulty of the movements themselves. Prior works have quantified skills by recognizing prescribed ``difficult'' body movements \cite{tang2020uncertainty, yu2021group, gao2020asymmetric} with the goal of automating scoring for solo sports. Skills underlying the sequence of body movements and their coordination and flexibility depending on the surrounding context, cannot be quantified by just recognizing each of the constituent actions.

How can we quantify such skills? Although this question clearly exceeds the scope of a single paper, we attempt to make a large stride by focusing on skills in dyadic human interactions. We examine a two-player sports, where skills surface and yield consequences within a limited timeframe more vividly than in daily life. In a dyadic sports, even though the repertoire of body movements is similar between the players, the nuances in execution and timing---conditioned on the game context including the opponent's moves---dictate the outcome. 

We select table tennis as our domain because matches can be captured effectively with a single camera, though our approach generalizes to other dyadic ball sports including tennis, badminton, and squash. Unlike prior visual modeling focusing on 3D ball and player analysis \cite{etaat2025latte, gossard2025tt3d}, our goal is to quantify intangible player skills.  We aim to quantify the player levels solely through observation of their plays---extracting an inherently latent quantity deeply encoded in the visual appearance of the players.

\begin{figure}[tb]
  \centering
  \includegraphics[width=\textwidth]{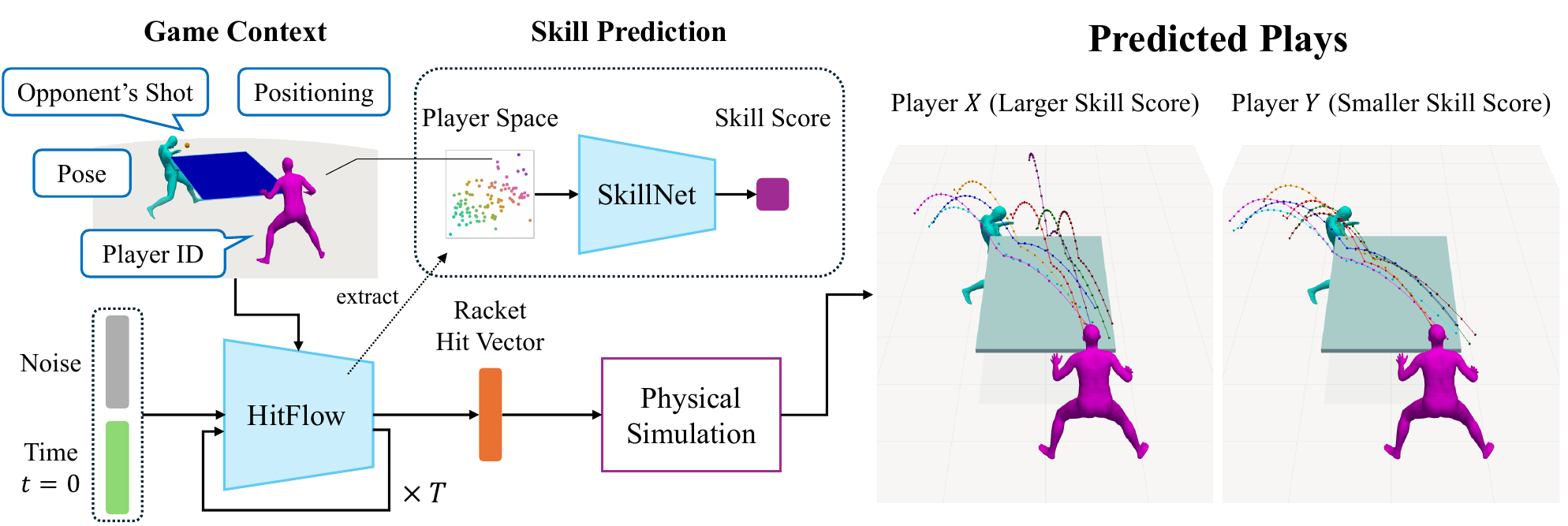}
  \caption{
    We model player semantics including their skill levels from observation of their plays. Our main model, HitFlow, jointly learns player embeddings and the distribution of racket hit vectors governing the ball trajectories in a play conditioned on the game context. The player embeddings encode semantics of each player including their skill levels. We demonstrate the extraction of such deeply latent characteristics of the players through probing and ranking of players and show their accuracy in match prediction.
  }
  \label{fig:overview}
\end{figure}

Our key idea is to learn a generative model that embodies each player's latent skills as reflected in their plays.
By learning a neural model that generates the possible moves of each player within a given game context, we can quantify skill levels.
We model these moves as the physical parameters of racket hits that cause the observed ball trajectories for each play. Given monocular videos, we first recover 2D ball trajectories, from which we recover the physical racket hit parameters using a Transformer-based model trained on a large-scale physics-simulation.
Once learned, these models provide player embeddings that encode latent tactics, from which we quantify the skill by estimating relative ranks using a RankNet-based framework.
We demonstrate the effectiveness of this approach through high accuracy in rank prediction for both known and unknown player groups. This novel framework for quantifying skills in dyadic sports provides a foundation for gauging skill sets in broader daily contexts.

\section{Related Works}
\paragraph{Action Analysis in Sports.}
Human action analysis in sports is an active area of research, with an increasing number of works focusing on dyadic ball sports.  
Etaat et al. \cite{etaat2025latte} introduce an autoregressive model to predict the future trajectory of a table tennis ball based on its past trajectory and the player's position and pose.
Kulkarni et al. \cite{kulkarni2021table} classify stroke types using temporal convolutional neural networks.
Ibh et al. \cite{ibh2023tempose} derive a Transformer-based method for player action recognition from spatio-temporal context, including the player's 2D joints and court positions, and the shuttlecock trajectories in badminton.
Although these studies address tasks such as action recognition and trajectory prediction, they do not attempt to analyze or quantify player skills.

Recent work by Ibh et al. \cite{ibh2024stroke} proposed a method for predicting the next stroke type based on 2D joints, court positions, and ID information of the players.
Although this approach leverages player embeddings to analyze distinct playing styles, it centers on categorical prediction. Our model transcends mere style analysis by learning a player space that encodes the rich, latent semantics regarding the skills of each player. 

\vspace{-8pt}
\paragraph{Action Quality Assessment.}
Prior research on action quality assessment revolve around classification, regression, and pairwise ranking estimation.
Liu et al. \cite{liu2021towards} estimate skill scores from surgical videos by integrating multiple features tailored to medical procedures. 
For competitive sports such as diving, Tang et al. \cite{tang2020uncertainty} perform score regression by modeling the uncertainty of ground-truth scores as a probability distribution.
To improve estimation accuracy, Yu et al. \cite{yu2021group} leverage reference data of identical actions to compute relative score differences, and Gao et al. \cite{gao2020asymmetric} incorporate interactions with other people or objects involved in the action. 
Although these approaches effectively quantify human skills in specific actions, they heavily rely on explicit supervision. This requirement for hand-labeled scores presents a significant bottleneck, as quantifying skill on an absolute scale is often non-trivial even for human experts. 

Zia et al. \cite{zia2016automated} address skill level classification in surgical videos, and further extend to include multi-modal accelerometer data \cite{zia2018video}.
Although these automate surgical skill assessment, they fundamentally rely on supervised class labels which are not only difficult to obtain but also inherently subjective and inconsistent.

Pairwise ranking offers an alternative for skill assessment. Bertasius et al. \cite{bertasius2017baller} employ ranking for basketball, though their model relies on domain-specific features that limit its generalizability to other sports. Doughty et al. \cite{doughty2017s} leverage a Siamese network to rank skills across diverse activities, from surgery to pizza dough rolling. Although this approach offers broad applicability, it is designed to discern skills at the video-instance level rather than the individual level. In contrast, our goal is the quantification of individual skills, which necessitates a holistic evaluation of players' cumulative activities rather than isolated clips.

\vspace{-8pt}
\paragraph{Skill Acquisition by Reinforcement Learning.}
Skill acquisition through reinforcement learning (RL) can also serve as a surrogate task for quantifying a person's skill set.
D’Ambrosio et al. \cite{dambrosio2025achieving} developed a robot system capable of defeating humans in table tennis by combining physics-based simulation with RL. By repeating matches against specific opponents, the robot identifies and exploits return patterns that the opponent struggles with. Although this captures strategic aspects of gameplay, the learned policy is inherently tailored to the robotic hardware, making it difficult to adapt for human players who cannot be easily simulated.
Other works have focused on modeling human-like agents in simulation. 
Zhang et al. \cite{yuan2023learning} extracted player actions from single-view tennis broadcasts to train humanoid agents. Although these agents can sustain rallies, they rely on predefined target return positions and do not explicitly model scoring strategies or quantitatively assess skill.
Similarly, Su et al. \cite{su2025hitter} developed a humanoid capable of returning ping-pong balls by incorporating human motion imitation. Its objective, however, remains limited to maintaining a rally. In contrast, our approach directly learns a generative representation of a player's unique behavior, enabling the quantification of skill and individual characteristics.

\section{Quantifying Table Tennis Skill Levels}
Our objective is to quantify player skill levels directly from competitive table tennis matches. To achieve this, our framework models individual player actions as a conditional probability distribution over the physical parameters of racket strokes, \ie, racket hits, that generate the plays in their games. To obtain the physical parameters of these racket hits for real gameplays, we train a neural model, HitFormer, to estimate the hit vector from observed 2D ball trajectories. The individual players can then be represented by the statistical distribution of these hit vectors in gameplays. We propose HitFlow, a generative model based on Flow Matching \cite{2022flow}, to represent this complex distribution conditioned on the surrounding game context. Simultaneously, we introduce a latent player space and learn individual embeddings that encode rich semantic information, capturing each player's unique characteristics and skill levels. We probe this player space to extract and quantify skill levels.

\subsection{Representing Plays With Racket Hits} \label{subsec:ball's_physics}
A trajectory of ping-pong ball shot by a player is the physical manifestation of their stroke and encapsulates their strategic decision-making at the specific instant. This suggests that the latent player characteristics including proficiency and skill levels, are intrinsically reflected in the statisical distribution of these ball trajectories over the course of gameplay. 

The ball trajectory can be modeled as a sequence of two physical states: flight and collision. In-air flight is dictated by gravity, drag, and Magnus force \cite{wang2019studies}
\begin{align}
m \dot{\bm{v}} = -k_{\mathrm{D}} \| \bm{v} \| \bm{v} + k_{\mathrm{M}} \bm{ \omega } \times \bm{v} + m \bm{g}\,, \label{eq:equation_of_motion}
\end{align}
where $m$ represents the mass of the ball, $ \bm{v} $ is the velocity, $ \bm{\omega} $ is the angular velocity, $ \bm{g} $ is the gravitational acceleration, and $ k_{\mathrm{D}} $ and $ k_{\mathrm{M}} $ are the drag and Magnus coefficients, respectively. The Magnus force, generated by the ball's spin, acts in the direction of the cross product of the angular velocity and velocity.

Collisions---occurring at the racket hit and table bounce---are modeled using the Coulomb's friction model \cite{bao2012bouncing},
\begin{align}
\bm{v}^{+} &= \bm{A} \bm{v}^{-} + \bm{B} \bm{ \omega }^{-}, \label{eq:bounce_v} \\
\bm{ \omega }^{+} &= \bm{C} \bm{v}^{-} + \bm{D} \bm{ \omega }^{-}\,, \label{eq:bounce_w}
\end{align}
where $\bm{v}^+$, $\bm{v}^-$, $\bm{\omega}^+$, and $\bm{\omega}^-$ represent the velocity and angular velocity immediately after and before the collision, respectively. The transition matrices $ \bm{A} $, $ \bm{B} $, $ \bm{C} $, and $ \bm{D} $ are determined by the state of the ball at impact. They account for the change between rolling and sliding, a transition governed by the coefficient $\alpha$: 
\begin{equation}
\alpha = \frac{\mu \left( 1+k_{\mathrm{COR}}|v_{z}^{-}| \right)}{\sqrt{\left(v_{x}^{-}-w_{y}^{-}r\right)^2 + \left(v_{y}^{-}-w_{x}^{-}r\right)^2}}\,, \label{eq:alpha}
\end{equation}
where $r$ represents the ball radius, $k_{\mathrm{COR}}$ denotes the coefficient of restitution, and $\mu$ is the coefficient of friction.

\cref{eq:equation_of_motion,eq:bounce_v,eq:bounce_w,eq:alpha} demonstrate that the entire trajectory of the ball is uniquely determined once the initial position $\bm{r}$, velocity $\bm{v}$, and angular velocity $\bm{\omega}$ are specified. Given this deterministic property, we define a hit vector as the concatenation of these parameters $\bm{r}$, $\bm{v}$, and $\bm{\omega}$ to serve as a compact representation of the ball's full trajectory. 

In turn, an observed trajectory practically determines the underlying hit vector, allowing for its recovery through inverse estimation. We devise a Transformer-based \cite{vaswani2017attention} architecture, which we refer to as HitFormer, to estimate the hit vector from observed 2D trajectories. HitFormer's self-attention mechanism ensures robust estimation even when player occlusions result in sparse or fragmented ball trajectory observations. To better capture the underlying geometric structure, 2D trajectories are first converted into Pl\"{u}cker coordinates using camera parameters before being fed into HitFormer, which subsequently outputs the refined 3D trajectories and hit vectors.

Even when estimation errors are small, significant bifurcations can occur depending on whether the ball bounces on the table edge. This distinction leads to vastly different outcomes, such as determining whether a shot is ``in'' or ``out.'' To address cases where the reprojection error exceeds a threshold due to these boundary effects, we apply an optimization step that directly minimizes the error. Since the initial estimates are generally reasonable, they can be efficiently refined with only a few iterations. This optimization phase ensures that our reconstructed trajectories are physically reliable and contextually accurate.

To train HitFormer, we generate SynthHit, a large-scale synthetic dataset of ping-pong ball trajectories containing approximately 3.5 million pairs of hit vectors and their corresponding trajectories. These trajectories are generated using physical simulation of \cref{eq:equation_of_motion,eq:bounce_v,eq:bounce_w,eq:alpha} and cover a comprehensive range of strokes, including smashes, drives, and chops. Once trained on SynthHit, HitFormer can accurately infer the hit vector from noisy 2D trajectory observations.

\subsection{Encoding Game Context} \label{subsec:reconstruct_players}
To incorporate the state of the match into our generative model, we must quantify the game context, which we define as the concurrent 3D motion and positioning of both competing players. Although this could include factors such as paddle poses and subtle wrist motions, we limit its scope to clearly observable visual cues in single-view videos. 

Our reconstruction pipeline leverages established pose estimation models integrated with the known geometry of the table. The player poses are first estimated as SMPL \cite{bogo2016keep} parameters using 4D-Humans \cite{goel2023humans}. By using the known geometry in world coordinate and the camera parameters estimated with the Perspective-n-Point (PnP) algorithm applied to the table corners, we derive a conversion function that maps the relative depth $d$ from monocular depth estimation \cite{chen2025video} to absolute distance $z$. 
We use the mask output from the SMPL parameters estimator \cite{goel2023humans} to identify the regions corresponding to players. By applying the ICP algorithm \cite{besl1992method} and by optimizing the SMPL mesh locations, we obtain the 3D world coordinates of the players.

\subsection{Modeling The Player} \label{subsec:player_model}
In any competitive table tennis scenario, a player's action is not uniquely determined. Actions are inherently stochastic, emerging from the complex interplay between the ball's trajectory, the opponent's movement, and the player's own physical and internal state. Attempting to predict a single, deterministic action ignores this fundamental stochasticity. 
To capture this structure, we leverage Flow Matching \cite{2022flow}, a generative model capable of capturing the continuous probabilistic distribution of data. By training a Flow Matching model to generate hit vectors, we effectively encode the player's actions including its stochasticity. 

\begin{figure}[tb]
  \centering
  \includegraphics[width=\textwidth]{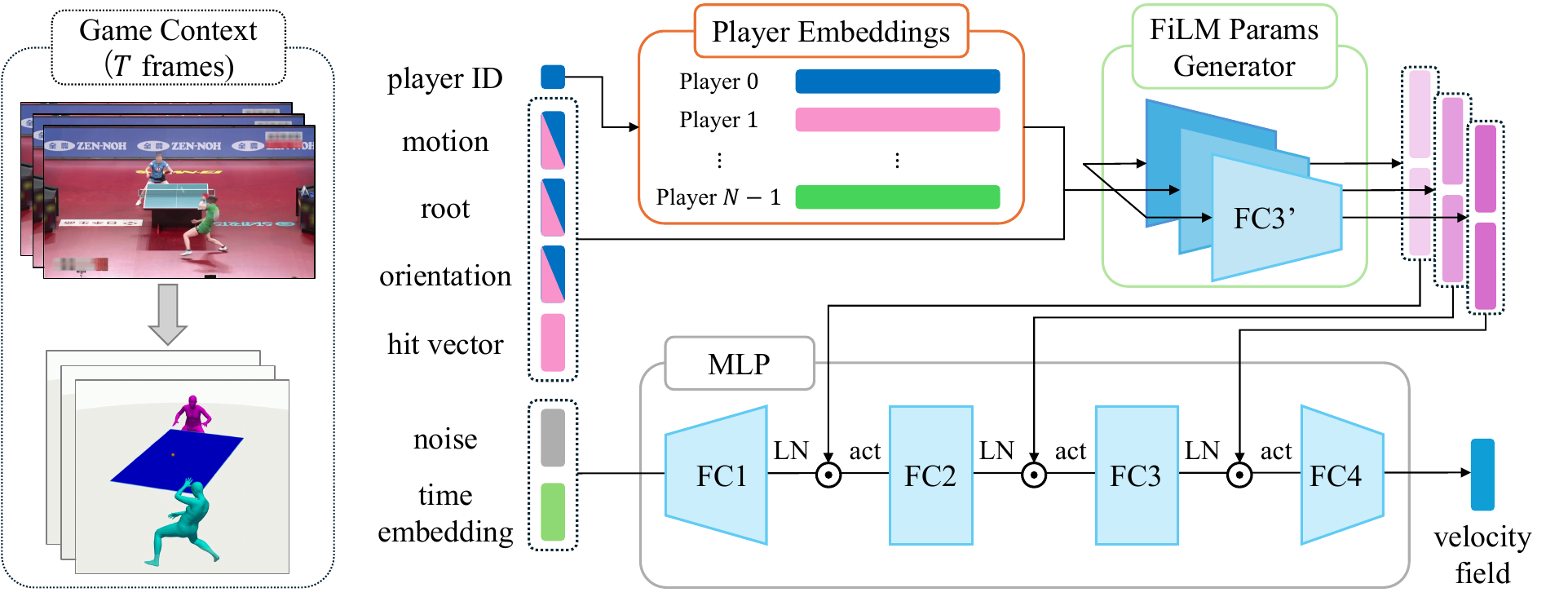}
  \caption{
    The architecture of HitFlow. The velocity field estimation by Flow Matching is conditioned on the game context and player identity through FiLM \cite{perez2018film}. Player embeddings are jointly learned with the model resulting in a player space encoding their latent characteristics including skill levels.
  }
  \label{fig:model_architecture}
\end{figure}

Players do not act with complete freedom; they must constantly adapt to the dynamic environment. The generative process must be conditioned on this game context.
We propose HitFlow, a generative model based on Flow Matching designed to capture the probabilistic characteristics of player actions conditioned on the state of the match.
HitFlow, as illustrated in \cref{fig:model_architecture}, takes a noise vector and a time embedding as inputs and outputs a velocity field with an MLP.
The velocity field estimation is conditioned on the game context via FiLM \cite{perez2018film}. FiLM enables the model to estimate conditioned velocity fields by applying feature-wise affine transformations to intermediate features, where the modulation parameters are generated from condition vectors. We define the game context as the motion, location, and orientation of both players over the past $T$ frames, as well as the opponent's preceding hit vector.

Even with an identical game context, the distribution of hit vectors varies significantly depending on the individual player, reflecting the skill level, \ie, exactly what we are seeking to model. To faithfully capture these player-specific characteristics, we introduce player embeddings which are learned jointly with the model. The MLP backbone of HitFlow learns fundamental physics of hit vector distributions, and the player embeddings and FiLM parameter generator capture the semantic information and stylistic nuances unique to each individual. 

Player motion is encoded using a pre-trained motion encoder \cite{zhu2023motionbert}. These motion features, along with location, orientation, and hit vectors, are  concatenated with player embeddings and input to the FiLM parameter generator. The resulting FiLM parameters $\bm{\beta}$ and $\bm{\gamma}$, modulate the intermediate representations normalized by LayerNorm \cite{ba2016layer}. Learning the model and player embeddings simultaneously enables HitFlow to predict velocity fields based on both the situational context and the identity of the player. The player space spanned by these learned embeddings encode player styles and tendencies, allowing us to quantify individual differences and achieve accurate context-aware hit vector prediction.

Once HitFlow is trained, it can generate possible trajectories conditioned on the game context. For instance, it allows prediction of the hit vector distribution of one player for a game context encountered by another player, by just swapping their embeddings. This capability allows us to visualize and intuitively understand the tactical differences between players as variations in generated trajectories, as illustrated in \cref{fig:overview}. 

\subsection{Ranking Players By Skill Levels}
The player embeddings learned by HitFlow (\cref{subsec:player_model}) capture rich semantic information of an individual's playing characteristics. We hypothesize that these embeddings inherently encode information related to a player's proficiency and strategic competence, \ie, skill level. To surface this latent skill level and establish a quantitative hierarchy, we employ RankNet \cite{burges2005learning}. 

RankNet is a pairwise ranking framework that maps a pair of inputs, $\mathbf{v}_i$ and $\mathbf{v}_j$, to a set of scalar scores, $s_i$ and $s_j$, respectively. The model is trained using relative ranking supervision (\eg, $i \succ j$, denoting that player $i$ possesses a higher skill level than player $j$) with a softplus loss function  
\begin{align}
\mathcal{L} &= \mathrm{softplus} \left( - \left( s_i - s_j \right) \right)\,. \label{eq:loss_function_rank}
\end{align}
We let RankNet process a pair of player embeddings as input to output scores that represent their relative skill levels. By aggregating these pairwise comparisons across the dataset, we can derive a global skill ranking for all players. 

\section{JTTA Table Tennis Dataset}
To analyze the strategic decision-making in table tennis and to quantify the player skills, we must observe players in high-stakes, dynamic matches where the skill influences action selection. Our model requires data that captures natural behaviors during competitive play. Unfortunately, there are no publicly available datasets of table tennis matches with player identities and corresponding statistics that can be used as ground truth, for instance, of rankings. 
For this study, we use official match footage from the All Japan Table Tennis Championships videos, provided by the Japan Table Tennis Association (JTTA), with their permission.

Since most high-level matches are recorded from single-view perspectives, we develop a pipeline to extract 3D physical and behavioral information from monocular broadcast videos. In table tennis, the stochastic interplay between players is dictated by the ball's physical dynamics. Because ball trajectories are highly sensitive to spin, it is essential to incorporate angular velocity into our analysis. For this, we curated a dataset that integrates the ball's physical state (spin and velocity), the players' articulated motions, and their absolute spatial coordinates in a unified 3D world coordinate frame.

We segment the raw match footage into individual rally clips using manual annotations for start and end frames. 
2D ball trajectories with a combination of TrackNetV3 \cite{chen2023tracknetv3} and manual refinement provide high-fidelity inputs for HitFormer (\cref{subsec:ball's_physics}). We define the world coordinate system based on the standardized geometry of the table. Camera intrinsic and extrinsic parameters are calculated via the PnP algorithm, using the table corners.

\vspace{-8pt}
\paragraph{3D Game Reconstruction.}
To reconstruct the hit vectors that govern each shot within a rally, we use HitFormer. Each shot's 2D trajectory is processed by HitFormer to estimate the underlying hit vectors. These hit vectors are then reprojected in the 2D image plane with physical simulation of the 3D trajectory. For instances where the reprojection error exceeds a predetermined threshold, we perform gradient-based optimization on hit vector, using the error as the loss function. This optimization is particularly crucial near the table edges, where small variations in hit parameters significantly alter the bounce dynamics.
The HitFormer is trained on the SynthHit dataset using random masking, which enables the model to be robust against occlusions even when critical moments, such as the impact or bounce, are missing from the observation. To ensure data reliability, estimated hit vectors are excluded from the experiments if their reprojection errors exceed a predefined threshold after the optimization process.

Players are reconstructed in 3D following \cref{subsec:reconstruct_players}. Floor region is determined with a pre-trained model \cite{ranftl2021vision}. The table top region is determined by projecting the table top in the world coordinate onto the image plane. Regions where players occlude the table top are excluded by subtracting the players' mask.

The resulting dataset comprises 36 matches from the women’s division and 34 from the men’s division. The source videos were recorded at 30 FPS with resolutions of 1920$\times$1080 or 1280$\times$720. After segmentation, the dataset contains approximately 1.08 million frames across 6,489 unique rallies, featuring 25 male and 28 female professional players. This scale ensures sufficient diversity in playstyles and skill levels for robust model training. The processed data will be released upon publication. 

\section{Experimental Results}

\subsection{Racket Hit Parameters}
\begin{table}[tb]
  \caption{3D ball trajectory reconstruction accuracy in Mean Absolute Error (MAE) in centimeters for various viewpoints. HitFormer consistently outperforms TT3D, particularly from the back viewpoint. Hit-vector optimization further refines accuracy.}
  \centering
  \setlength{\tabcolsep}{10pt}
  \begin{tabular}{@{}lccc@{}}
    \toprule
    method & side & oblique & back \\
    \midrule
    TT3D & 12.4 & 17.1 & 29.8 \\
    HitFormer (w/o opt.) & \textbf{10.3} & \textbf{16.3} & \textbf{18.8} \\
    HitFormer (w/ opt.) & \textbf{10.1} & \textbf{16.2} & \textbf{18.5} \\
  \bottomrule
  \end{tabular}
  \label{tab:tt3d_experiment}
\end{table}
We first validate the accuracy of 3D trajectory reconstruction performed by HitFormer. For this, we can use TT3D \cite{gossard2025tt3d} dataset, a benchmark dataset that provides ground-truth 3D ball trajectories, corresponding 2D observations, and camera parameters.
We evaluate the performance of HitFormer in terms of Mean Absolute Error (MAE) and compare against the optimization-based recovery method introduced in TT3D.

As shown in \cref{tab:tt3d_experiment}, HitFormer demonstrates higher accuracy across all viewpoints. Notably, the error from the back viewpoint---which typically suffers from significant depth degeneracy---is substantially lower than the baseline. This suggests that HitFormer effectively leverages the physical priors learned from our SynthHit dataset to resolve ambiguities that purely optimization-based methods cannot handle. Data in TT3D are captured in 25 FPS, which is smaller than that of the videos in our dataset. Although capturing data with larger frame rate may improve reconstruction accuracy, these results demonstrate that it is possible to reconstruct 3D trajectories at regular camera frame rates. 

\subsection{Hit Vector Distribution}
We evaluate HitFlow using energy score \cite{gneiting2007strictly} to verify whether it learns hit vector distribution appropriately. We assess the validity of hit vector distribution through trajectory distribution, which is induced from the distribution.
The energy score is a metric which captures a trade-off between the prediction's discrepancy from ground truth and the diversity of predictions, and defined as
\begin{equation}
    \varepsilon = \mathbb{E} \left[ d\left(X, y\right) \right] - \frac{1}{2} \mathbb{E} \left[ d\left(X,X'\right) \right] \,, \label{eq:energy_score}
\end{equation}
where $X$ and $X'$ are independent samples from the predicted distribution, $y$ is the observed ground truth sample, and $d$ is a distance function. We adopt MAE as the distance function.

We compare HitFlow with LATTE-MV, a Transformer-based autoregressive baseline.
The model requires a temporal sequence of the opponent's 3D SMPL joint positions, own root position, and ball position, which are covered by our dataset.
The model is designed to estimate probabilistic regions containing the ball via ensemble and conformal prediction.
To enable a direct comparison using the energy score, we train an ensemble of deterministic LATTE-MV models using (i) the full training dataset with five different random seeds (seed ensemble) and (ii) five independent training sub-dataset (partitioned ensemble). For HitFlow, which is inherently stochastic, we sample 100 trajectories to ensure statistical reliability of the metric. Unlike the ensemble-based LATTE-MV, which requires training multiple models, HitFlow can efficiently generate a large number of samples without additional training or significant computational overhead.

\begin{figure}[tb]
  \centering
  \includegraphics[width=\linewidth]{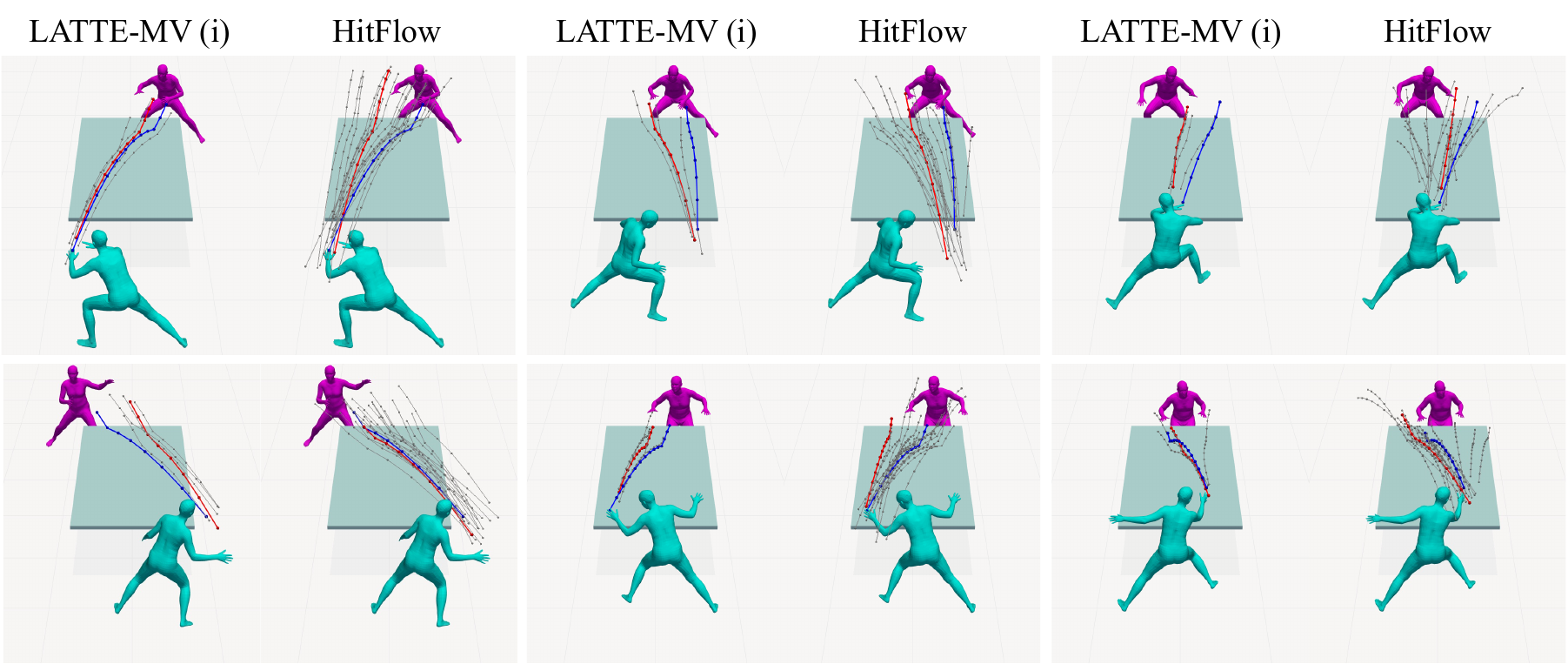}
  \caption{
    Hit vector (ball trajectory) prediction. For each pair, the left shows LATTE-MV prediction (seed ensemble) and the right shows HitFlow prediction. The GT trajectories are shown in blue and the trajectories where the MAE takes median in the samples are shown in red as representatives. HitFlow outputs diverse trajectories and covers GT within the distribution.
  }
  \label{fig:latte_mv_experiment}
\end{figure}

Both models are trained on our dataset due to the availability of hit vectors as part of the game context. These hit vectors can also be instantiated as 3D ball trajectories with physical simulation, which can be used for training LATTE-MV. We provide the game context to both models, ending two frames before the target shot. LATTE-MV model uses the coordinates of the ego player's dominant hand as input, but due to the absence of this information in our dataset, we apply zero padding instead.
\cref{tab:latte_mv_experiment} shows that HitFlow outperforms both ensemble types of LATTE-MV in energy score. \cref{fig:latte_mv_experiment} demonstrates the diversity of HitFlow outputs, which covers the ground-truth trajectory.

\begin{table}[tb]
  \caption{Ball trajectory prediction accuracy comparison. HitFlow outperforms LATTE-MV for both ensemble settings.}
  \centering
  \setlength{\tabcolsep}{5pt}
  \begin{tabular}{@{}lccc@{}}
    \toprule
    method & HitFlow & LATTE-MV (i) & LATTE-MV (ii) \\
    \midrule
    energy score (↓) & \textbf{0.37} & 0.40 & 0.40 \\
  \bottomrule
  \end{tabular}
  \label{tab:latte_mv_experiment}
\end{table}

\subsection{Player Space Analysis}
Player embeddings are expected to encode semantic information of individual players including tangible characteristics latent to skill levels. To verify whether the player embeddings truly contain such information, we performed linear probing of the space they span for sex, handedness, height and age classification. Additionally, we qualitatively examined whether player embeddings reflect play styles by applying t-SNE \cite{maaten2008visualizing} to the player space.

\vspace{-8pt}
\subsubsection{Downstream Classification Performance.}
The player space is expected to be structured in order to reflect players' semantic information to hit vector generation efficiently. We evaluate how readily the information is decodable from the player space using linear probing. We give labels to 6-30 selected embeddings and train a linear classifier on the player embeddings. 
We compute Matthews Correlation Coefficient (MCC) to examine the classification performance. An MCC of $+1$ corresponds to a perfect classification, $0$ to a random classification, and $-1$ to a completely inversed classification. As classification labels, we select male or female, right-handed or left-handed, older or younger than the median age, and taller or shorter than the median height. 

\cref{fig:linear_probing} demonstrates that the player space reflect the differences in sex, handedness, and height, but do not contain information regarding age. These results are reasonable; gender difference often manifests as physical strength difference, which is one of the key elements that characterizes a player. Handedness difference affects the tendency of return directions, and height difference affects reach range and footwork patterns. Age difference might become apparent when we consider a wider range of players in age, but it does not come to the surface in this experiment. Although we focused on players born from the late 80s to the late 2000s, matches for analysis are held between 2022 and 2025 and players are competing on the same stage within the same period. This likely causes no significant disparity across age.

\vspace{-8pt}
\subsubsection{Semantic Similarity Analysis.}
To examine whether similar players lie close to each other in our learned player space, we focus on three players with distinct characteristics, namely female and chopper (player \textsf{14}, \textsf{15}, and \textsf{43}) in our dataset. The play style of a chopper is so distinct from others that it can act as a landmark in the embedding space. We apply t-SNE \cite{maaten2008visualizing} to the player space, using cosine distance for the computation.

\begin{figure}[tb]
    \centering
    \begin{subfigure}[c]{0.49\textwidth}
        \centering
        \includegraphics[width=\textwidth]{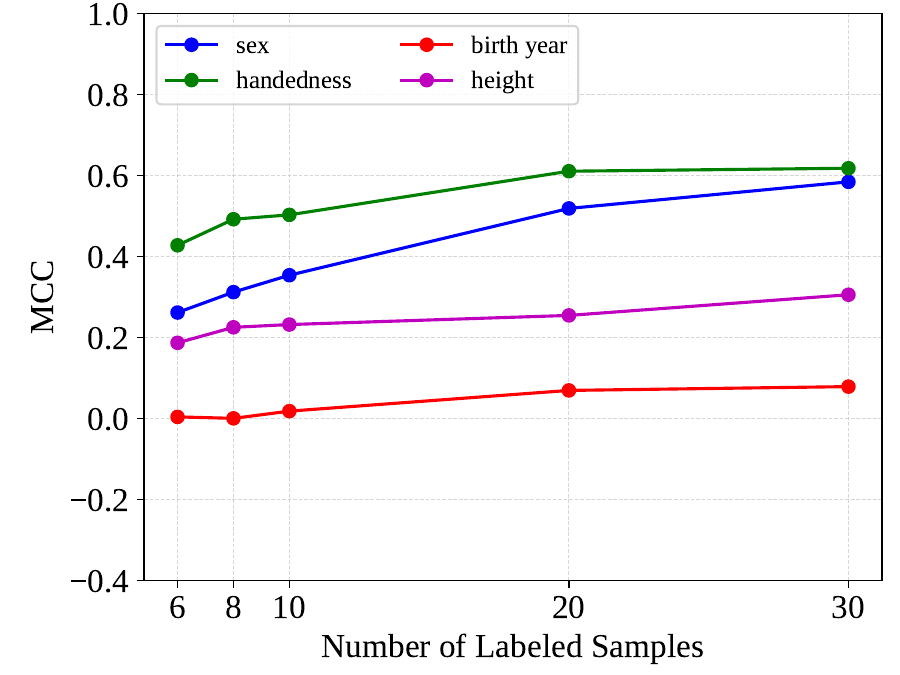}
        \caption{MCC of linear probing.}
        \label{fig:linear_probing}
    \end{subfigure}
    \hfill
    \begin{subfigure}[c]{0.49\textwidth}
        \centering
        \includegraphics[width=\textwidth]{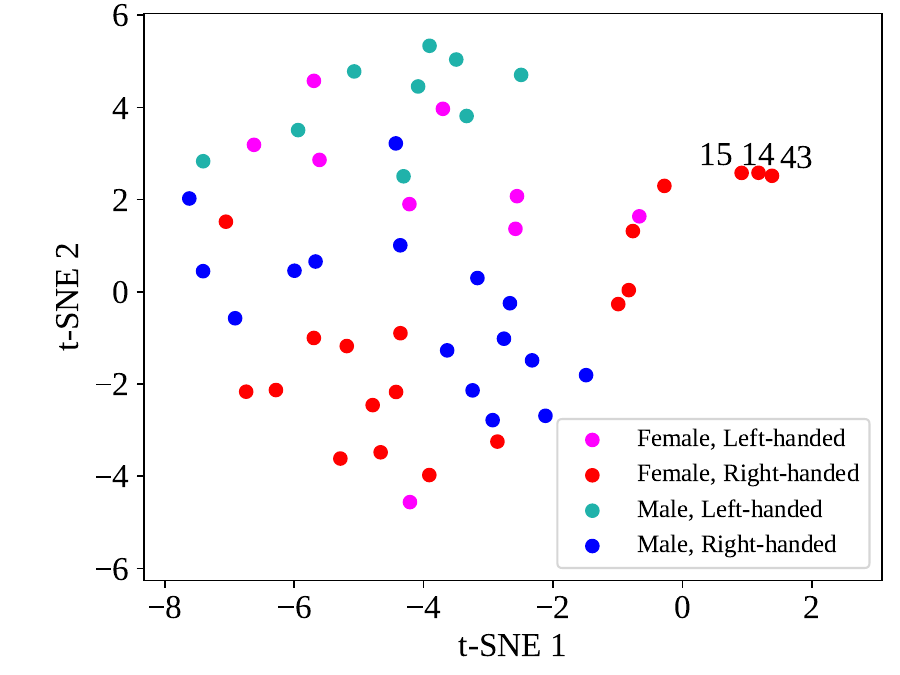}
        \caption{Player space map by t-SNE.}
        \label{fig:cosine_distance}
    \end{subfigure}
    \caption{
        Matthew Correlation Coefficient (MCC) (\emph{left}) and t-SNE result for the player space by cosine distance.
        (a) MCC is sufficiently large even with a small number of labels, and tend to increase as the number of labels grow for sex, handedness, and height. Age appears to be unrelated to the player embeddings.
        (b) Player \textsf{14}, \textsf{15} and \textsf{43}, who are all female choppers, are particularly close to each other. Sex and handedness are reflected in the distributional bias simultaneously, showing that the embeddings encode sex, handedness and play style reflected in the observed plays. 
    }
\end{figure}

\cref{fig:cosine_distance} shows the t-SNE result. Player \textsf{14}, \textsf{15} and \textsf{43} are distributed in a cluster away from the main group. Sex and handedness are also reflected in the distribution, showing that the embeddings encode sex, handedness, and play style reflected in the observed plays.

\subsection{Player Skill Level Estimation}
To extract skill levels from the learned player space, we employ RankNet \cite{burges2005learning} and assess whether the embeddings truly capture this information. We design a small MLP, which we call SkillNet, with the assumption that the player embeddings contain information pertaining to the levels of skills of each player so clearly that it is possible for even a small MLP to capture it, and to avoid overfitting due to the small number of player embeddings, 53. Although direct labels for skill levels are unavailable, we have labels for rank, which can be regarded as a proxy. We perform rank prediction, where players' skill level are quantified as the score SkillNet outputs, and use it as a proxy evaluation of how well the player space encodes the varying skill levels. We define rank labels as tournament results of the All Japan Table Tennis Championships and also the players' world rankings.

\vspace{-8pt}
\subsubsection{Ranking Estimation in Known Group} \label{subsubsec:rank_estimation_known}
To assess the accuracy of skill level estimation, we can predict an unknown player's rank within a group of known players.
For this, we can randomly select one player in the $N$ players and train SkillNet on the remaining $N-1$ player embeddings to output scalar scores representing relative ranks. Higher scores indicate higher skill levels. 
Using this trained SkillNet, the held-out player's rank can be determined among the other $N-1$ players. By repeating this process with different held-out embeddings independently, we obtain $N$ predicted ranks. We compare these predicted ranks with the true ranks and evaluate the accuracy of the prediction with Spearman's correlation coefficient and its $p$-value with the significance level of 5 \%.

\begin{figure}[t]
\centering
\begin{minipage}[t]{0.49\textwidth}
    \centering
    \begin{subfigure}[c]{0.49\textwidth}
        \centering
        \includegraphics[width=\textwidth]{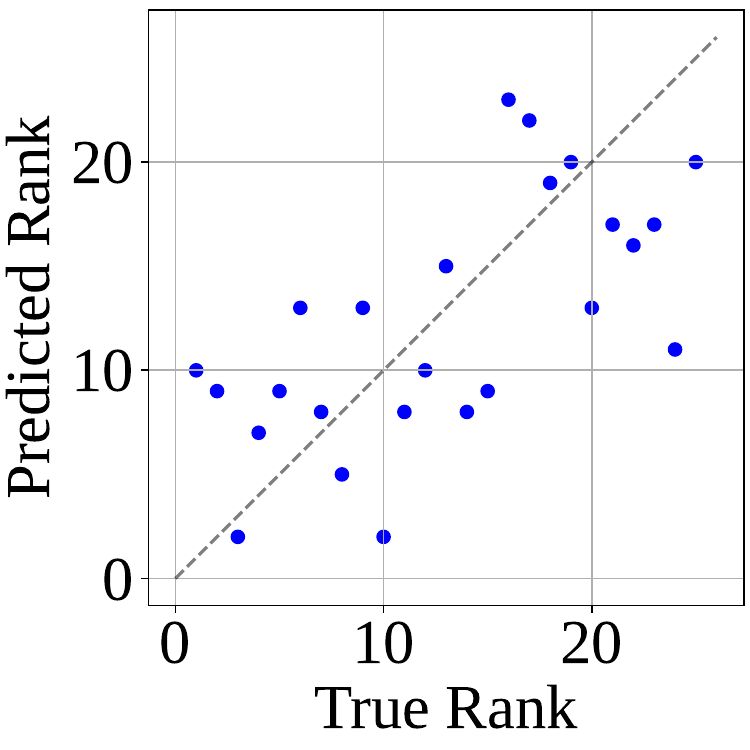}
        \caption{Corr. for male.}
        \label{fig:rank_estimation_known_M}
    \end{subfigure}
    \hfill
    \begin{subfigure}[c]{0.49\textwidth}
        \centering
        \includegraphics[width=\textwidth]{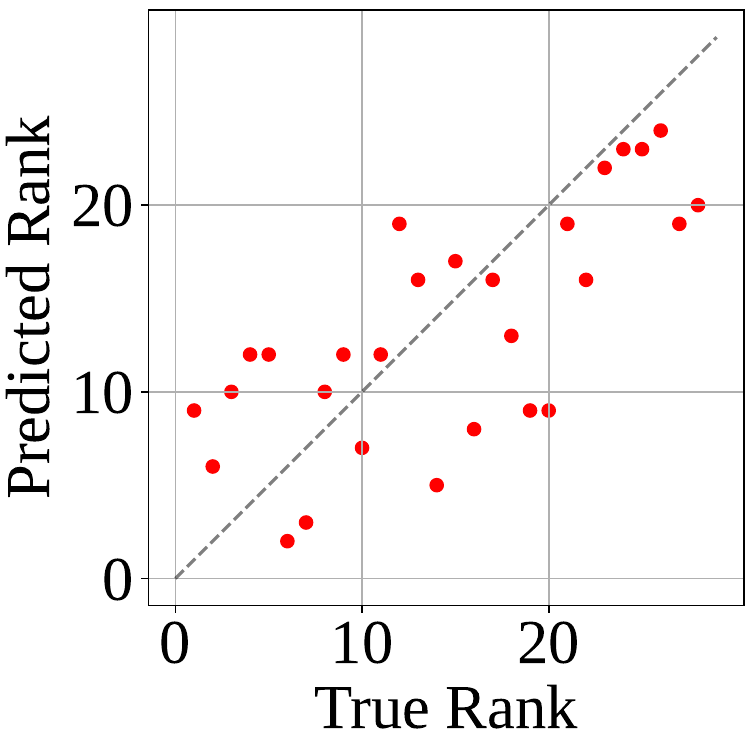}
        \caption{Corr. for female.}
        \label{fig:rank_estimation_known_F}
    \end{subfigure}
    \caption{
        Correlation between predicted and true ranks (a) for male and (b) for female. Significant correlation between predicted and true ranks for both male and female players can be observed.
    }
    \label{fig:rank_estimation_known}
\end{minipage}\hfill
\begin{minipage}[t]{0.49\textwidth}
    \centering
    \begin{subfigure}[c]{0.49\textwidth}
        \centering
        \includegraphics[width=\textwidth]{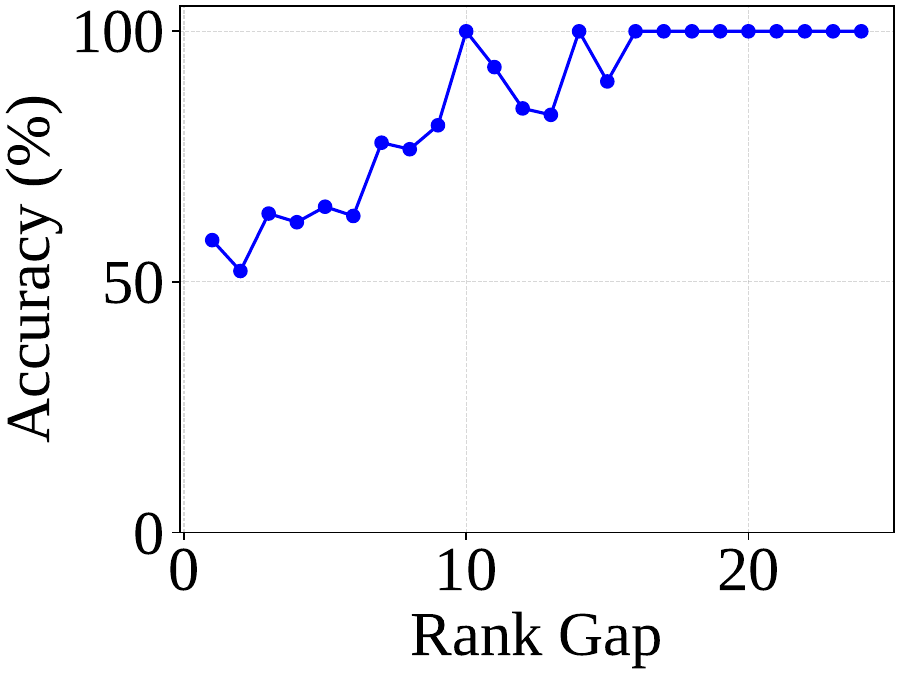}
        \caption{Acc. for male.}
        \label{fig:rank_estimation_unknown_M}
    \end{subfigure}\hfill
    \begin{subfigure}[c]{0.49\textwidth}
        \centering
        \includegraphics[width=\textwidth]{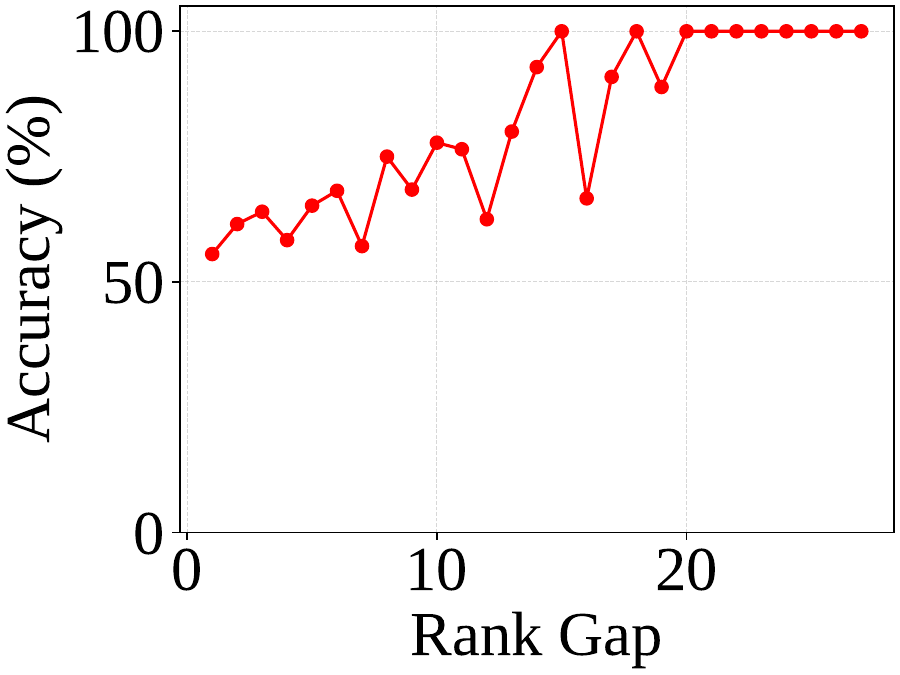}
        \caption{Acc. for female.}
        \label{fig:rank_estimation_unknown_F}
    \end{subfigure}
    \caption{
        Accuracies of relative rank estimation between two unknown players (a) for male and (b) for female. For both male and female, accuracy is relatively small when the rank gap is small and for bigger rank gaps the prediction accuracy increases.
    }
    \label{fig:rank_estimation_unknown}
\end{minipage}

\end{figure}
\cref{fig:rank_estimation_known} shows that there is a significant correlation between the predicted and the true ranks for both male and female players. The correlation coefficient is 0.665 ($p<0.001$) for male and 0.699 ($p<0.001$) for female. This suggests that the learned player space contain rich information related to skill levels and we can successfully probe it with a simple MLP.

Since SkillNet is trained using rankings derived from the same matches HitFlow learns, the model might implicitly memorize match-specific features rather than generalized skill. To verify the model's robustness against such data leakage, we evaluated its performance using official world rankings, which are independent of our dataset's match outcomes. We observed significant correlations of 0.652 ($p < 0.001$) for males and 0.578 ($p = 0.003$) for females, confirming that the model captures generalized proficiency. The slight reduction in the female correlation is likely due to limited data; four players without official world rankings were excluded, reducing the training set by approximately 100 relative rank labels.

\vspace{-8pt}
\subsubsection{Ranking Estimation between Unknown Players} \label{subsubsec:rank_estimation_unknown}
We can also gauge how well the skill levels are encoded in the learned player space by predicting relative ranks between two unknown players.
For this task, we randomly select 2 players and train SkillNet with the remaining $N-2$ players. We then infer the scores for the two held-out players by feeding their player embeddings into the trained SkillNet. The output scores let us determine the relative superiority of the two players. By repeating this process for all combinations of held-out players, we can assess the accuracy of the predictions which reflects the accuracy of skill levels encoded in the player embeddings.

\begin{table}[tb]
  \caption{Accuracy of rank prediction between unknown players. Overall accuracy for both male and female is over 70 \%, and when the rank gap becomes larger than 5, the accuracy becomes higher.}
  \centering
  \setlength{\tabcolsep}{10pt}
  \begin{tabular}{@{}lccc@{}}
    \toprule
    sex & overall & rank gap $>5$ \\
    \midrule
    Male & 77.3 \% & 87.4 \% \\
    Female & 73.5 \% & 79.8 \% \\
  \bottomrule
  \end{tabular}
  \label{tab:rank_estimation_unknown}
\end{table}
As shown in \cref{fig:rank_estimation_unknown}, the accuracy of predicted relative skill levels increases as the ground-truth rank gap between two players widens. Detailed results for both the overall accuracy and cases where the rank difference exceeds 5 are summarized in \cref{tab:rank_estimation_unknown}. These findings indicate that while it is straightforward to rank players with disparate skill levels, it is challenging to do so for closely ranked players. In high-level competition, such as the national-level championships in our dataset, matches between closely ranked opponents are often so competitive that a clear hierarchy is difficult to define even for human experts. This difficulty underscores our hypothesis that skill is encoded in subtle nuances rather than linear metrics. Nevertheless, our results demonstrate that the learned player space captures the fundamental characteristics of these skill levels. While actual match prediction would require real-time gameplay interpretation, our model and resulting player embeddings provide a robust, player-specific prior for such tasks.

\section{Conclusion}
Towards quantifying human skill level, we focused on table tennis where skills surface more vividly than actions in daily life.
By representing game plays with racket hits and by modeling them for each player with a stochastic generative model, we showed that a player space can be jointly learned which encodes the player characteristics underlying individual skill levels. By probing this learned player space, we demonstrated that skill levels can be quantified for ranking predictions. We believe this novel approach to human skill quantification can be applied to dyadic human interactions beyond competitive sports, which we hope to explore in future work. 
Immediate future work towards this goal include better accuracy of 3D reconstruction and increasing the player pool for larger-scale analysis. Both will be essential to extend our work beyond sports as finer and broader modeling of observed interactions would likely directly enrich the jointly learned embedding space of individuals. 

\section*{Acknowledgements}
We would like to express our sincere gratitude to Japan Table Tennis Association (JTTA) and SAN-EI Corporation for their generous cooperation in providing the data used in this study. This work was in part supported by JSPS KAKENHI 21H04893 and JST JPMJAP2305.

%
%
\bibliographystyle{splncs04}
\bibliography{main}

@String(CVPR  = {IEEE Conf. Comput. Vis. Pattern Recog.})

@String(ICCV  = {Int. Conf. Comput. Vis.})

@String(ECCV  = {Eur. Conf. Comput. Vis.})

@String(ICML  = {Int. Conf. Mach. Learn.})

@String(ICLR  = {Int. Conf. Learn. Represent.})

@String(CVPRW = {IEEE Conf. Comput. Vis. Pattern Recog. Worksh.})

@String(AAAI  = {AAAI})

@String(CVPR  = {CVPR})

@String(ICCV  = {ICCV})

@String(ECCV  = {ECCV})

@String(ICML  = {ICML})

@String(ICLR  = {ICLR})

@String(CVPRW = {CVPRW})

@inproceedings{kulkarni2021table,
  title={Table {T}ennis {S}troke {R}ecognition {U}sing {T}wo-{D}imensional {H}uman {P}ose {E}stimation},
  author={Kulkarni, Kaustubh Milind and Shenoy, Sucheth},
  booktitle=CVPRW,
  pages={4576--4584},
  year={2021}
}

@inproceedings{ibh2024stroke,
  title={A {S}troke of {G}enius: {P}redicting the {N}ext {M}ove in {B}adminton},
  author={Ibh, Magnus and Gra{\ss}hof, Stella and Hansen, Dan Witzner},
  booktitle=CVPRW,
  pages={3376--3385},
  year={2024}
}

@inproceedings{ibh2023tempose,
  title={TemPose: {A} {N}ew {S}keleton-{B}ased {T}ransformer {M}odel {D}esigned for {F}ine-{G}rained {M}otion {R}ecognition in {B}adminton},
  author={Ibh, Magnus and Grasshof, Stella and Witzner, Dan and Madeleine, Pascal},
  booktitle=CVPRW,
  pages={5199--5208},
  year={2023}
}

@inproceedings{liu2021towards,
  title={Towards {U}nified {S}urgical {S}kill {A}ssessment},
  author={Liu, Daochang and Li, Qiyue and Jiang, Tingting and Wang, Yizhou and Miao, Rulin and Shan, Fei and Li, Ziyu},
  booktitle=CVPR,
  pages={9522--9531},
  year={2021}
}

@inproceedings{tang2020uncertainty,
  title={Uncertainty-aware {S}core {D}istribution {L}earning for {A}ction {Q}uality {A}ssessment},
  author={Tang, Yansong and Ni, Zanlin and Zhou, Jiahuan and Zhang, Danyang and Lu, Jiwen and Wu, Ying and Zhou, Jie},
  booktitle=CVPR,
  pages={9839--9848},
  year={2020}
}

@inproceedings{yu2021group,
  title={Group-{A}ware {C}ontrastive {R}egression for {A}ction {Q}uality {A}ssessment},
  author={Yu, Xumin and Rao, Yongming and Zhao, Wenliang and Lu, Jiwen and Zhou, Jie},
  booktitle=ICCV,
  pages={7919--7928},
  year={2021}
}

@inproceedings{gao2020asymmetric,
  title={An {A}symmetric {M}odeling for {A}ction {A}ssessment},
  author={Gao, Jibin and Zheng, Wei-Shi and Pan, Jia-Hui and Gao, Chengying and Wang, Yaowei and Zeng, Wei and Lai, Jianhuang},
  booktitle=ECCV,
  pages={222--238},
  year={2020},
  organization={Springer}
}

@article{zia2016automated,
  title={Automated {V}ideo-{B}ased {A}ssessment of {S}urgical {S}kills for {T}raining and {E}valuation in {M}edical {S}chools},
  author={Zia, Aneeq and Sharma, Yachna and Bettadapura, Vinay and Sarin, Eric L and Ploetz, Thomas and Clements, Mark A and Essa, Irfan},
  journal={International journal of computer assisted radiology and surgery},
  volume={11},
  number={9},
  pages={1623--1636},
  year={2016},
  publisher={Springer}
}

@article{zia2018video,
  title={Video and {A}ccelerometer-{B}ased {M}otion {A}nalysis for {A}utomated {S}urgical {S}kills {A}ssessment},
  author={Zia, Aneeq and Sharma, Yachna and Bettadapura, Vinay and Sarin, Eric L and Essa, Irfan},
  journal={International journal of computer assisted radiology and surgery},
  volume={13},
  number={3},
  pages={443--455},
  year={2018},
  publisher={Springer}
}

@inproceedings{doughty2017s,
    author={Doughty, Hazel and Damen, Dima and Mayol-Cuevas, Walterio},
    title={Who’s {B}etter? {W}ho’s {B}est? {P}airwise {D}eep {R}anking for {S}kill {D}etermination},
    booktitle=CVPR,
    year={2017}
}

@inproceedings{bertasius2017baller,
  title={Am {I} a {B}aller? {B}asketball {P}erformance {A}ssessment from {F}irst-{P}erson {V}ideos},
  author={Bertasius, Gedas and Soo Park, Hyun and Yu, Stella X and Shi, Jianbo},
  booktitle=ICCV,
  pages={2177--2185},
  year={2017}
}

@inproceedings{bao2012bouncing,
  title={Bouncing {M}odel for the {T}able {T}ennis {T}rajectory {P}rediction and the {S}trategy of {H}itting the {B}all},
  author={Bao, Han and Chen, Xiaopeng and Wang, Zhan Tao and Pan, Min and Meng, Fei},
  booktitle={2012 IEEE International Conference on Mechatronics and Automation},
  pages={2002--2006},
  year={2012},
  organization={IEEE}
}

@article{wang2019studies,
  title={Studies and simulations on the flight trajectories of spinning table tennis ball via high-speed camera vision tracking system},
  author={Wang, Ping and Zhang, Qian and Jin, Yinli and Ru, Feng},
  journal={Proceedings of the Institution of Mechanical Engineers, Part P: Journal of Sports Engineering and Technology},
  volume={233},
  number={2},
  pages={210--226},
  year={2019},
  publisher={SAGE Publications Sage UK: London, England}
}

@article{yuan2023learning,
  title={Learning {P}hysically {S}imulated {T}ennis {S}kills from {B}roadcast {V}ideos},
  author={Yuan, Ye and Makoviychuk, Viktor and Guo, Y and Fidler, S and Peng, X and Fatahalian, K},
  journal={ACM Trans. Graph},
  volume={42},
  number={4},
  year={2023}
}

@inproceedings{dambrosio2025achieving,
  title={Achieving {H}uman {L}evel {C}ompetitive {R}obot {T}able {T}ennis},
  author={D'Ambrosio, David B and Abeyruwan, Saminda and Graesser, Laura and Iscen, Atil and Amor, Heni Ben and Bewley, Alex and Reed, Barney J and Reymann, Krista and Takayama, Leila and Tassa, Yuval and others},
  booktitle={2025 IEEE International Conference on Robotics and Automation (ICRA)},
  pages={74--82},
  year={2025},
  organization={IEEE}
}

@article{su2025hitter,
  title={{HITTER}: {A} {H}umano{I}d {T}able {TE}nnis {R}obot via {H}ierarchical {P}lanning and {L}earning},
  author={Su, Zhi and Zhang, Bike and Rahmanian, Nima and Gao, Yuman and Liao, Qiayuan and Regan, Caitlin and Sreenath, Koushil and Sastry, S Shankar},
  journal={arXiv preprint arXiv:2508.21043},
  year={2025}
}

@inproceedings{bogo2016keep,
  title={Keep it {SMPL}: {A}utomatic {E}stimation of 3{D} {H}uman {P}ose and {S}hape from a {S}ingle {I}mage},
  author={Bogo, Federica and Kanazawa, Angjoo and Lassner, Christoph and Gehler, Peter and Romero, Javier and Black, Michael J},
  booktitle=ECCV,
  pages={561--578},
  year={2016},
  organization={Springer}
}

@inproceedings{chen2025video,
  title={Video {D}epth {A}nything: {C}onsistent {D}epth {E}stimation for {S}uper-{L}ong {V}ideos},
  author={Chen, Sili and Guo, Hengkai and Zhu, Shengnan and Zhang, Feihu and Huang, Zilong and Feng, Jiashi and Kang, Bingyi},
  booktitle=CVPR,
  pages={22831--22840},
  year={2025}
}

@inproceedings{goel2023humans,
  title={Humans in 4{D}: {R}econstructing and {T}racking {H}umans with {T}ransformers},
  author={Goel, Shubham and Pavlakos, Georgios and Rajasegaran, Jathushan and Kanazawa, Angjoo and Malik, Jitendra},
  booktitle=ICCV,
  pages={14783--14794},
  year={2023}
}

@inproceedings{besl1992method,
  title={Method for {R}egistration of {3-D} {S}hapes},
  author={Besl, Paul J and McKay, Neil D},
  booktitle={Sensor fusion IV: control paradigms and data structures},
  volume={1611},
  pages={586--606},
  year={1992},
  organization={Spie}
}

@inproceedings{gossard2025tt3d,
  title={{TT3D}: {T}able {T}ennis 3{D} {R}econstruction},
  author={Gossard, Thomas and Ziegler, Andreas and Zell, Andreas},
  booktitle=CVPRW,
  pages={5821--5831},
  year={2025}
}

@inproceedings{etaat2025latte,
  title={{LATTE-MV}: {L}earning to {A}nticipate {T}able {T}ennis {H}its from {M}onocular {V}ideos},
  author={Etaat, Daniel and Kalaria, Dvij and Rahmanian, Nima and Sastry, S Shankar},
  booktitle=CVPR,
  pages={7115--7124},
  year={2025}
}

@inproceedings{chen2023tracknetv3,
  title={Tracknetv3: {E}nhancing {S}huttlecock {T}racking with {A}ugmentations and {T}rajectory {R}ectification},
  author={Chen, Yu-Jou and Wang, Yu-Shuen},
  booktitle={Proceedings of the 5th ACM International Conference on Multimedia in Asia},
  pages={1--7},
  year={2023}
}

@inproceedings{ranftl2021vision,
  title={Vision {T}ransformers for {D}ense {P}rediction},
  author={Ranftl, Ren{\'e} and Bochkovskiy, Alexey and Koltun, Vladlen},
  booktitle=ICCV,
  pages={12179--12188},
  year={2021}
}

@inproceedings{perez2018film,
  title={{FiLM}: {V}isual {R}easoning with a {G}eneral {C}onditioning {L}ayer},
  author={Perez, Ethan and Strub, Florian and De Vries, Harm and Dumoulin, Vincent and Courville, Aaron},
  booktitle=AAAI,
  volume={32},
  number={1},
  year={2018}
}

@inproceedings{2022flow,
    author={Lipman, Yaron and Chen, Ricky TQ and Ben-Hamu, Heli and Nickel, Maximilian and Le, Matt},
    title={Flow {M}atching for {G}enerative {M}odeling},
    booktitle=ICLR,
    year=2023 
}

@article{vaswani2017attention,
  title={Attention is {A}ll {Y}ou {N}eed},
  author={Vaswani, Ashish and Shazeer, Noam and Parmar, Niki and Uszkoreit, Jakob and Jones, Llion and Gomez, Aidan N and Kaiser, {\L}ukasz and Polosukhin, Illia},
  journal={Advances in neural information processing systems},
  volume={30},
  year={2017}
}

@inproceedings{burges2005learning,
  title={Learning to {R}ank using {G}radient {D}escent},
  author={Burges, Chris and Shaked, Tal and Renshaw, Erin and Lazier, Ari and Deeds, Matt and Hamilton, Nicole and Hullender, Greg},
  booktitle=ICML,
  pages={89--96},
  year={2005}
}

@inproceedings{zhu2023motionbert,
  title={Motion{BERT}: A {U}nified {P}erspective on {L}earning {H}uman {M}otion {R}epresentations},
  author={Zhu, Wentao and Ma, Xiaoxuan and Liu, Zhaoyang and Liu, Libin and Wu, Wayne and Wang, Yizhou},
  booktitle=ICCV,
  pages={15085--15099},
  year={2023}
}

@article{maaten2008visualizing,
  title={Visualizing {D}ata using t-{SNE}},
  author={Maaten, Laurens van der and Hinton, Geoffrey},
  journal={Journal of machine learning research},
  volume={9},
  number={Nov},
  pages={2579--2605},
  year={2008}
}

@article{gneiting2007strictly,
  title={Strictly {P}roper {S}coring {R}ules, {P}rediction, and {E}stimation},
  author={Gneiting, Tilmann and Raftery, Adrian E},
  journal={Journal of the American statistical Association},
  volume={102},
  number={477},
  pages={359--378},
  year={2007},
  publisher={Taylor \& Francis}
}

@article{ba2016layer,
  title={Layer {N}ormalization},
  author={Ba, Jimmy Lei and Kiros, Jamie Ryan and Hinton, Geoffrey E},
  journal={arXiv preprint arXiv:1607.06450},
  year={2016}
}

@inproceedings{loshchilov2019decoupled,
  title={Decoupled {W}eight {D}ecay {R}egularization},
  author={Loshchilov, Ilya and Hutter, Frank},
  booktitle=ICLR,
  year={2019}
}

@inproceedings{loshchilov2016sgdr,
  title={{SGDR}: {S}tochastic {G}radient {D}escent with {W}arm {R}estarts},
  author={Loshchilov, Ilya and Hutter, Frank},
  booktitle=ICLR,
  year={2017}
}

@article{srivastava2014dropout,
  title={Dropout: {a} {S}imple {W}ay to {P}revent {N}eural {N}etworks from {O}verfitting},
  author={Srivastava, Nitish and Hinton, Geoffrey and Krizhevsky, Alex and Sutskever, Ilya and Salakhutdinov, Ruslan},
  journal={The journal of machine learning research},
  volume={15},
  number={1},
  pages={1929--1958},
  year={2014},
  publisher={JMLR. org}
}

@article{branch1999subspace,
  title={A {S}ubspace, {I}nterior, and {C}onjugate {G}radient {M}ethod for {L}arge-{S}cale {B}ound-{C}onstrained {M}inimization {P}roblems},
  author={Branch, Mary Ann and Coleman, Thomas F and Li, Yuying},
  journal={SIAM Journal on Scientific Computing},
  volume={21},
  number={1},
  pages={1--23},
  year={1999},
  publisher={SIAM}
}

\appendix

\section{Physics of Ping-pong Ball}
The transition matrices $ \bm{A} $, $ \bm{B} $, $ \bm{C} $, and $ \bm{D} $, determined by $\alpha$, map the translational and angular velocities immediately before impact to those immediately after impact.
When $\alpha$ is less than 0.4, sliding effects dominate and the matrices $\bm{A}$, $\bm{B}$, $\bm{C}$, and $\bm{D}$ are
\begin{equation}
\begin{alignedat}{2}
    \bm{A} &= 
        \begin{bmatrix}
        0.6 & 0 & 0 \\
        0 & 0.6 & 0 \\
        0 & 0 & -k_{\mathrm{COR}}
        \end{bmatrix}
    & \qquad
    \bm{B} &= 
        \begin{bmatrix}
        0 & 0.4r & 0 \\
        -0.4r & 0 & 0 \\
        0 & 0 & 0
        \end{bmatrix}
    \\[6pt]
    \bm{C} &= 
        \begin{bmatrix}
        0 & -\frac{0.6}{r} & 0 \\
        \frac{0.6}{r} & 0 & 0 \\
        0 & 0 & 0
        \end{bmatrix}
    & \qquad
    \bm{D} &= 
        \begin{bmatrix}
        0.4 & 0 & 0 \\
        0 & 0.4 & 0 \\
        0 & 0 & 1
        \end{bmatrix}\,.
\end{alignedat} \label{eq:matrices_slide}
\end{equation}
When $\alpha$ is larger than 0.4, rolling effects dominate and the matrices $\bm{A}$, $\bm{B}$, $\bm{C}$, and $\bm{D}$ are 
\begin{equation}
\begin{alignedat}{2}
    \bm{A} &= 
        \begin{bmatrix}
        1-\alpha & 0 & 0 \\
        0 & 1-\alpha & 0 \\
        0 & 0 & -k_{\mathrm{COR}}
        \end{bmatrix}
    & \qquad
    \bm{B} &= 
        \begin{bmatrix}
        0 & \alpha r & 0 \\
        -\alpha r & 0 & 0 \\
        0 & 0 & 0
        \end{bmatrix}
    \\[6pt]
    \bm{C} &= 
        \begin{bmatrix}
        0 & -\frac{3\alpha}{2r} & 0 \\
        \frac{3\alpha}{2r} & 0 & 0 \\
        0 & 0 & 0
        \end{bmatrix}
    & \qquad
    \bm{D} &= 
        \begin{bmatrix}
        1-\frac{3}{2}\alpha & 0 & 0 \\
        0 & 1-\frac{3}{2}\alpha & 0 \\
        0 & 0 & 1
        \end{bmatrix}\,.
\end{alignedat} \label{eq:matrices_roll}
\end{equation}

\section{Statistics of Dataset}
\subsection{SynthHit Dataset Generation}
The SynthHit dataset, used to train HitFormer, comprises a diverse array of hit vectors. We explicitly configured the parameter space to reflect the physics of ten distinct shot categories: banana flick, chop, drive, lob, serve, smash, push, other long, other short, and other. Additionally, we defined a uniform distribution across the union of these parameter spaces to sample ``random shots.'' These samples populate the intermediate regions between predefined categories, ensuring the model generalizes beyond discrete labels. Only ``valid'' trajectories---those that successfully land within the legal bounds of the table---are included in the final dataset.

\begin{figure}[tb]
  \centering
  \includegraphics[width=0.8\linewidth]{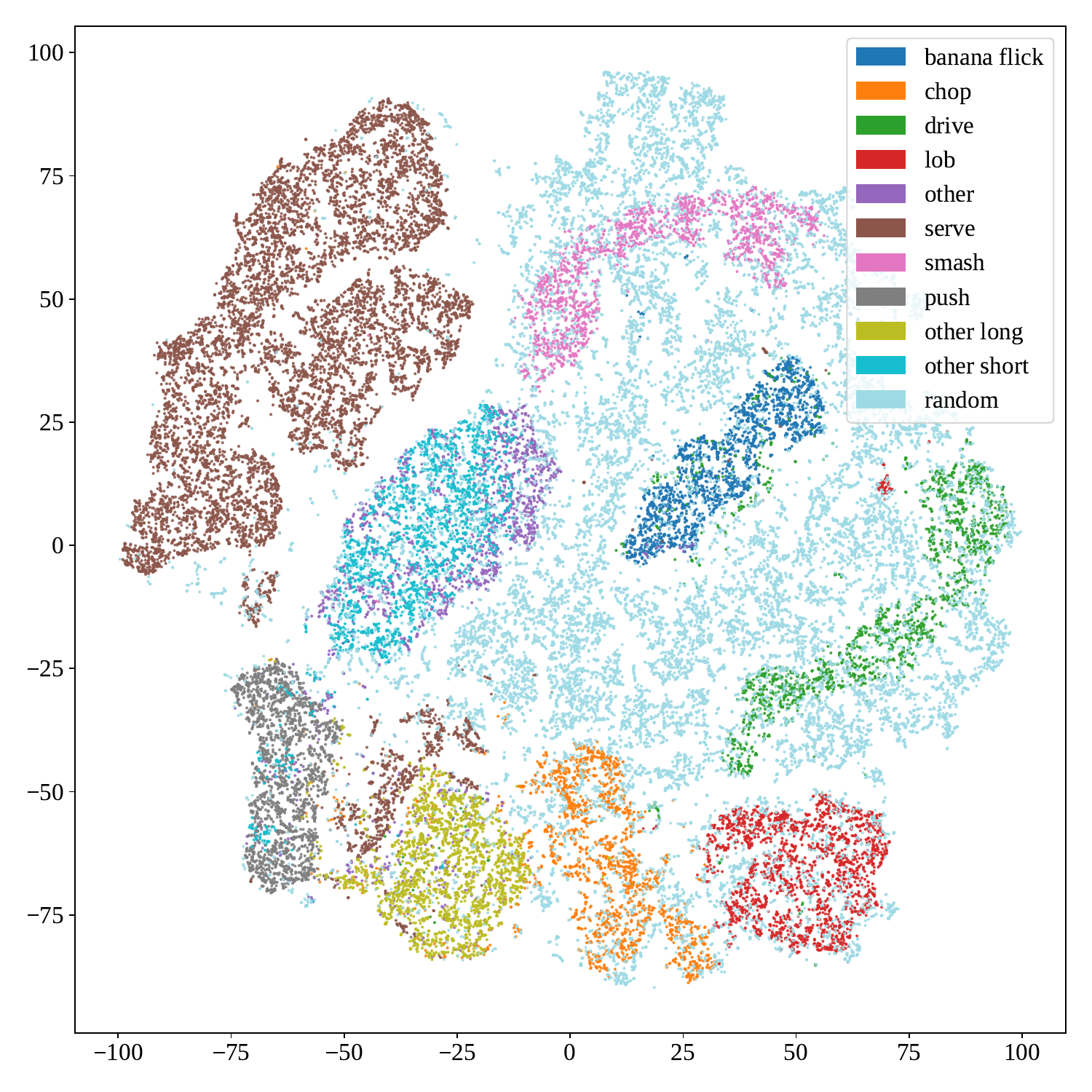}
  \caption{t-SNE visualization of hit vectors in SynthHit. Predefined shot types form distinct clusters, while random samples bridge the transition regions between them.}
  \label{fig:tsne_SynthHit}
\end{figure}

\cref{fig:tsne_SynthHit} provides a t-SNE \cite{maaten2008visualizing} visualization of the hit vectors in SynthHit. While most shot types form distinct clusters with random samples filling the transitions between them, the serve cluster is notably isolated. Unlike other strokes, serves must bounce first on the player’s own side, creating a unique physical signature that precludes intermediate trajectories between it and offensive strokes.

\subsection{JTTA Table Tennis Dataset}
\begin{figure}[tb]
  \centering
  \includegraphics[width=\linewidth]{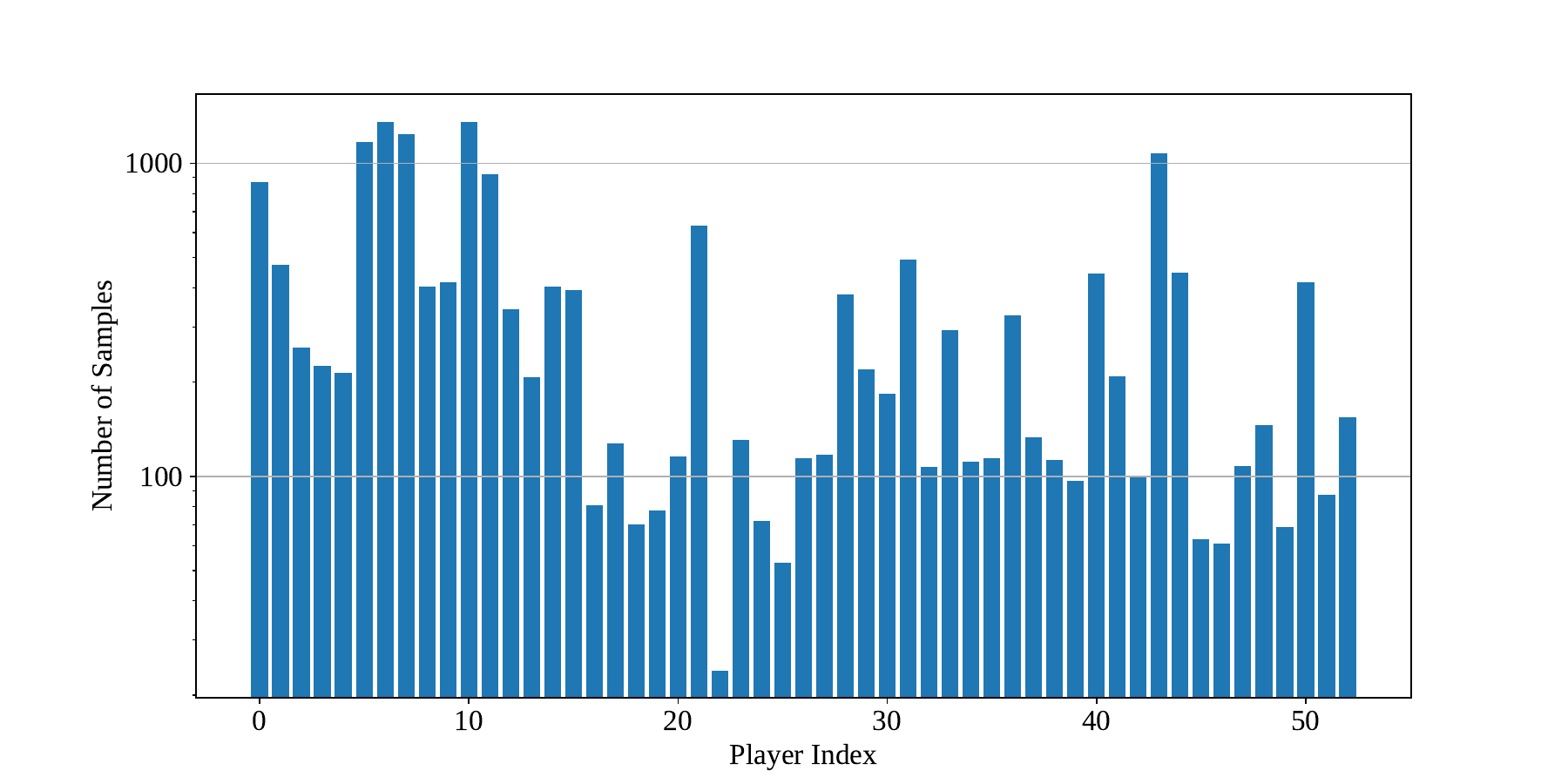}
  \caption{Number of rally samples available per player in the JTTA dataset.}
  \label{fig:jtta_each_player}
\end{figure}

The JTTA table tennis dataset includes trajectory and pose data for 53 professional players. \cref{fig:jtta_each_player} illustrates the distribution of samples per player.
Due to the nature of professional tournament structures, the dataset exhibits a natural imbalance: higher-skilled players appear more frequently. This is because elite players advance further in single-season brackets and tend to qualify for championships over multiple years, resulting in a higher volume of recorded match footage.

\section{Qualitative Effects of Optimization on Hit Vectors}
\begin{figure}[tb]
    \centering
    \begin{subfigure}[c]{0.49\textwidth}
        \centering
        \includegraphics[width=\textwidth]{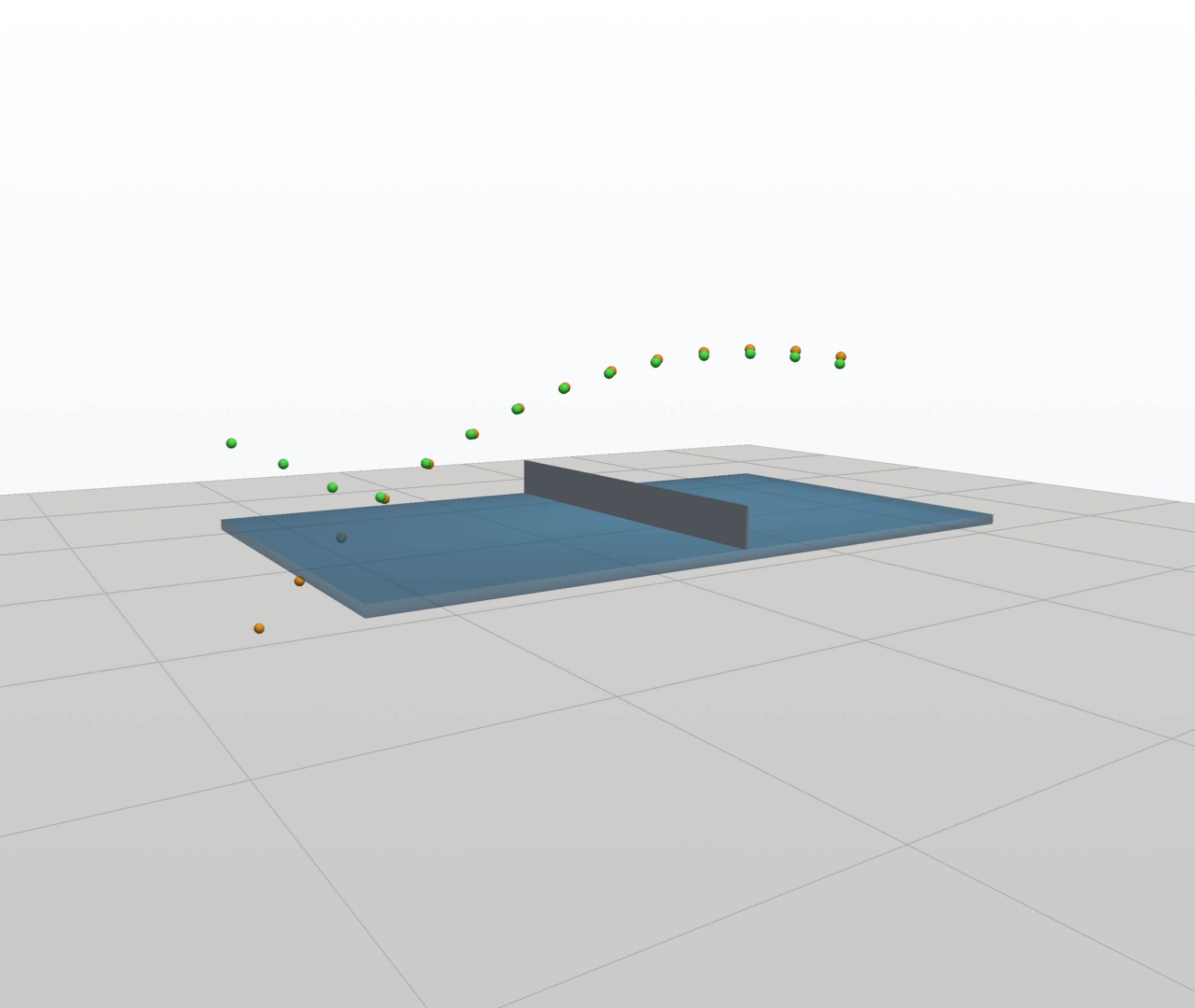}
        \caption{Before optimization.}
        \label{fig:before_optimization}
    \end{subfigure}
    \hfill
    \begin{subfigure}[c]{0.49\textwidth}
        \centering
        \includegraphics[width=\textwidth]{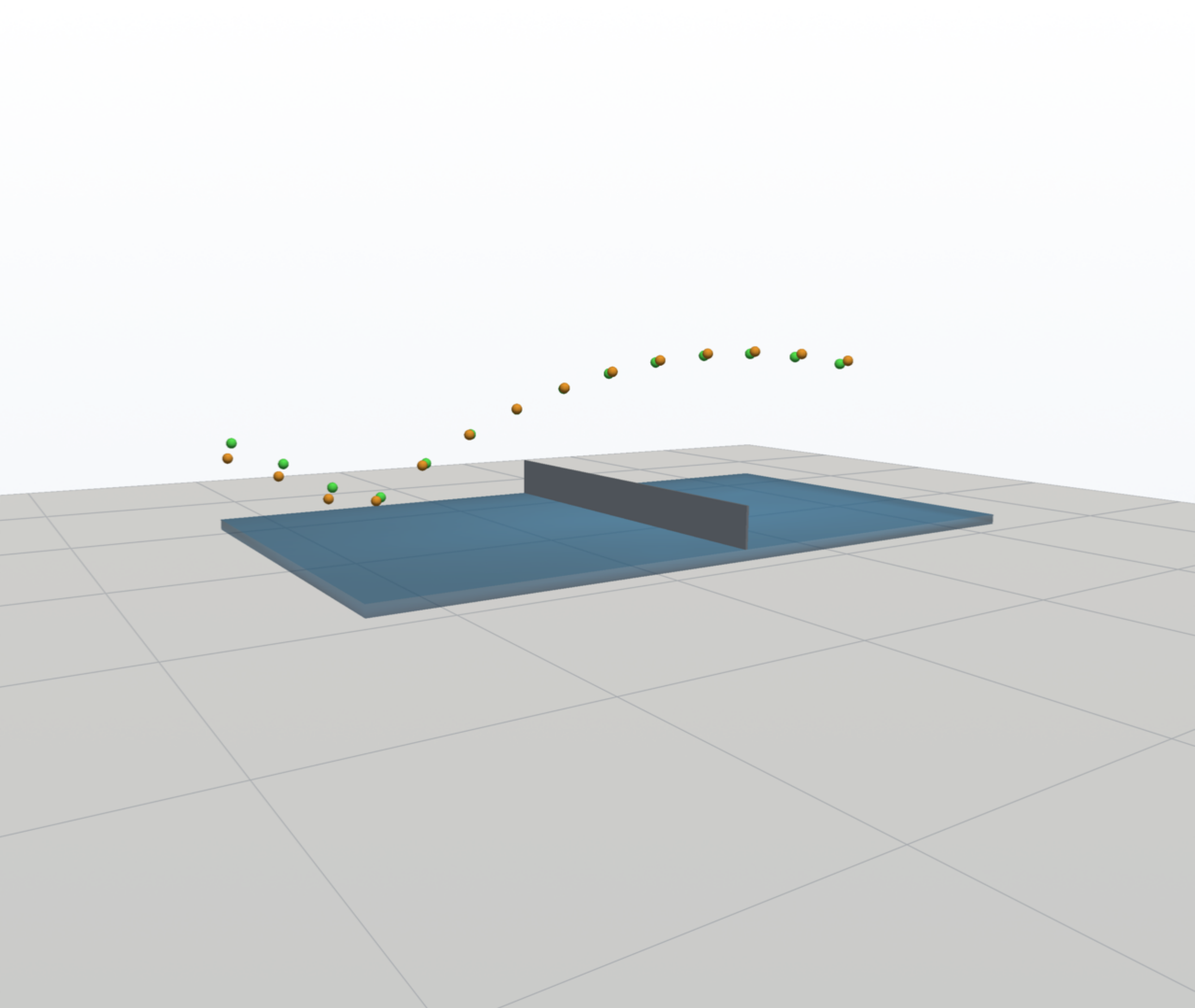}
        \caption{After optimization.}
        \label{fig:after_optimization}
    \end{subfigure}
    \caption{
Trajectory estimation from the observation viewpoint: (a) before and (b) after optimization. Green points denote ground truth, and orange points represent estimates. In the pre-optimized state, the estimation fails to capture the bounce discontinuity, resulting in a large reprojection error. Post-optimization, this deviation is resolved, accurately capturing the impact on the table.
    }
    \label{fig:effect_optimization}
\end{figure}

When reconstructing hit vectors from observed 2D trajectories, we employ an optimization process for cases where the initial reprojection error exceeds a predefined threshold. This refinement stage allows the estimation to recover from significant deviations, particularly those caused by the high sensitivity of the trajectory to bounce events. \cref{fig:effect_optimization} illustrates a representative case where the optimization successfully restores a physically plausible trajectory from an initially divergent estimate.
By minimizing the reprojection error, the optimization ensures that the final hit vectors are both physically reliable and visually consistent with the evidence. We implemented this refinement using the \texttt{least\_squares} function with trust region reflective algorithm \cite{branch1999subspace} from the SciPy library.

\section{Implementation Details}
To ensure robust training of HitFlow, we apply a multi-level dropout strategy \cite{srivastava2014dropout} to three core components: player embeddings, condition vectors, and the modulation mechanism. First, embedding-level dropout is applied by randomly replacing the entire player embedding with a zero vector. This prevents the model from treating the player space as a simple lookup table and forces it to learn generalized features. Second, we apply vector-level dropout to the condition signals; each constituent vector is independently replaced with a learnable null vector. This partial removal encourages the embeddings to capture player-specific information rather than allowing the model to rely solely on environmental context. Third, we apply modulation dropout to the FiLM mechanism \cite{perez2018film} by randomly bypassing the affine transformation. This encourages HitFlow to internalize fundamental generative concepts of the hit vector distribution. We set the dropout probabilities at 0.1 for player embeddings, 0.8 for condition vectors, and 0.6 for the modulation mechanism. These combined constraints prevent ``shortcut learning'' and ensure that the latent space remains semantically rich.

We use the AdamW optimizer \cite{loshchilov2019decoupled} paired with a CosineAnnealingScheduler \cite{loshchilov2016sgdr} for 3,000 epochs. The initial learning rate is set to $10^{-4}$ for the MLP and player embeddings, and $10^{-5}$ for the FiLM parameter generator, with all components decaying to a terminal rate of $10^{-6}$. To mitigate the data imbalance shown in \cref{fig:jtta_each_player}, we employ balanced sampling across players during training. The MLP hidden layer dimension is set to 128, player embeddings to 32, and the temporal window for game context is fixed at 10 frames.

\end{document}